\documentclass[sigconf]{acmart} 
\acmSubmissionID{509}

\usepackage{booktabs} 
\usepackage{enumitem}
\setitemize{noitemsep,topsep=0pt,parsep=0pt,partopsep=0pt}
\usepackage[ruled]{algorithm2e} 

\SetAlFnt{\small}
\SetAlCapFnt{\small}
\SetAlCapNameFnt{\small}
\SetAlCapHSkip{0pt}

\copyrightyear{2023}
\acmYear{2023}
\setcopyright{rightsretained}
\acmConference[SA Conference Papers '23]{SIGGRAPH Asia 2023 Conference Papers}{December 12--15, 2023}{Sydney, NSW, Australia}
\acmBooktitle{SIGGRAPH Asia 2023 Conference Papers (SA Conference Papers '23), December 12--15, 2023, Sydney, NSW, Australia}
\acmDOI{10.1145/3610548.3618185}
\acmISBN{979-8-4007-0315-7/23/12}

\begin{document}
\title{What is the Best Automated Metric for Text to Motion Generation?}

\author{Jordan Voas}
\orcid{0000-0002-4663-5680}
\affiliation{
  \institution{University of Texas at Austin}
  \streetaddress{2317 Speedway}
  \city{Austin}
  \state{TX}
  \postcode{78712}
  \country{USA}}
\email{jvoas@utexas.edu}
\author{Yili Wang}
\affiliation{
  \institution{University of Texas at Austin}
  \streetaddress{2317 Speedway}
  \city{Austin}
  \state{TX}
  \postcode{78712}
  \country{USA}}
\email{ywang98@utexas.edu}
\author{Qixing Huang}
\affiliation{
 \institution{University of Texas at Austin}
  \streetaddress{2317 Speedway}
  \city{Austin}
  \state{TX}
  \postcode{78712}
  \country{USA}}
\email{huangqx@cs.utexas.edu}
\author{Raymond Mooney}
\orcid{0000-0002-4504-0490}
\affiliation{
  \institution{University of Texas at Austin}
  \streetaddress{2317 Speedway}
  \city{Austin}
  \state{TX}
  \postcode{78712}
  \country{USA}}
\email{mooney@cs.utexas.edu}


\begin{teaserfigure}
\centering
\fcolorbox{gray}{white}{\includegraphics[width=0.95\linewidth, trim={0 0.5cm 0 27.5cm},clip]{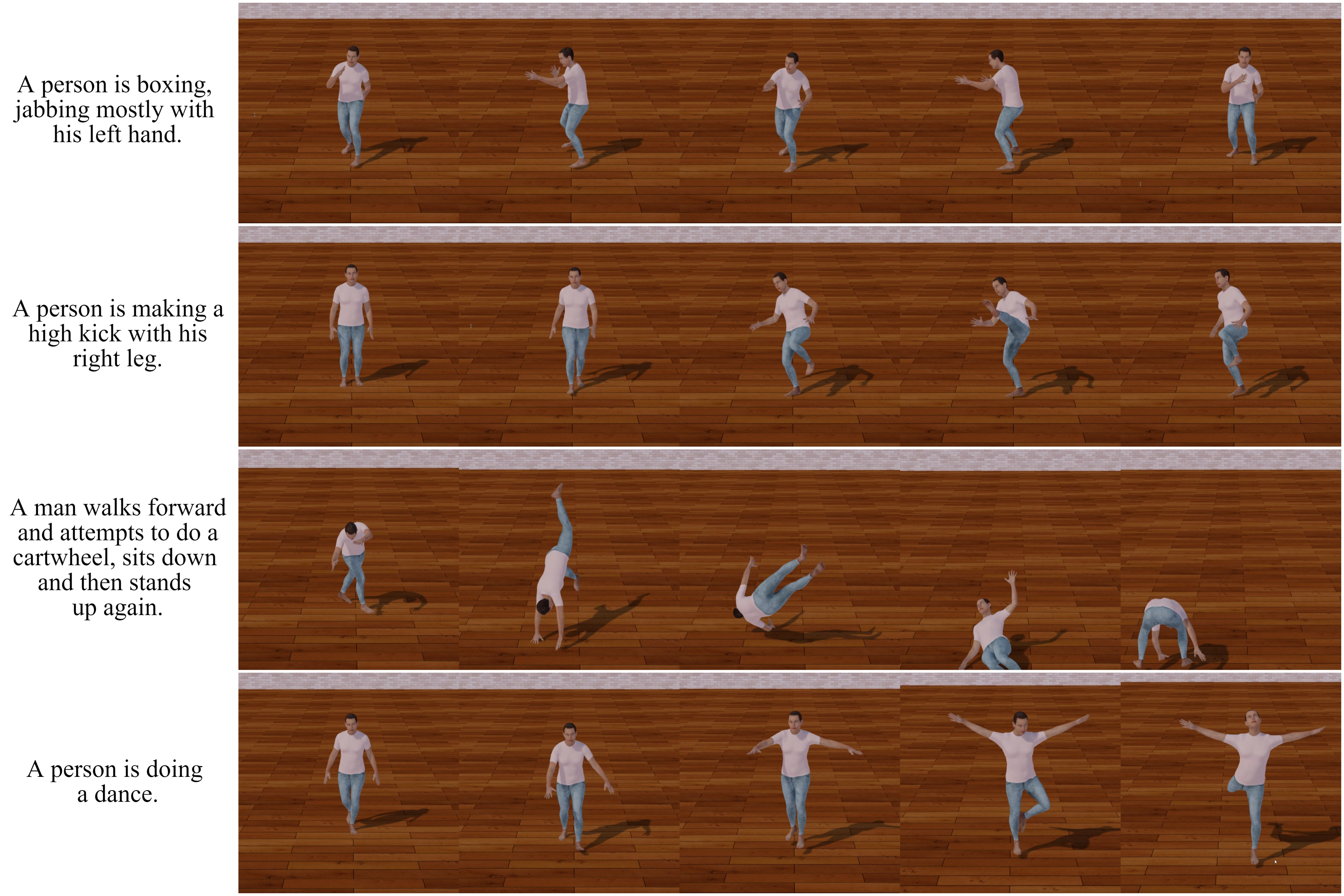}}
\caption{Sampled motion frames with paired descriptions, as used in our human evaluations. Our rendering framework generates pseudo-realistic environments with skin, wall, and floor textures as well as environment lighting and steady camera motions. }
\label{fig:motion-exps}
\end{teaserfigure}

\begin{abstract}
There is growing interest in generating skeleton-based human motions from natural language descriptions. While most efforts have focused on developing better neural architectures for this task, there has been no significant work on determining the proper evaluation metric. Human evaluation is the ultimate accuracy measure for this task, and automated metrics should correlate well with human quality judgments. Since descriptions are compatible with many motions, determining the right metric is critical for evaluating and designing effective generative models. This paper systematically studies which metrics best align with human evaluations and proposes new metrics that align even better. Our findings indicate that none of the metrics currently used for this task show even a moderate correlation with human judgments on a sample level. However, for assessing average model performance, commonly used metrics such as R-Precision and less-used coordinate errors show strong correlations. Additionally, several recently developed metrics are not recommended due to their low correlation compared to alternatives. We also introduce a novel metric based on a multimodal BERT-like model, MoBERT, which offers strongly human-correlated sample-level evaluations while maintaining near-perfect model-level correlation. Our results demonstrate that this new metric exhibits extensive benefits over all current alternatives.
\end{abstract}


%
%


\maketitle

\section{Introduction}

High-quality human motion generation in animation has a wide range of applications, from creating realistic CGI in cinema to enabling context-aware character movement in video games. The increasing interest in generating human motions from natural language descriptions (text-to-motion) is evident \cite{Lin2018GeneratingAV,DBLP:conf/3dim/AhujaM19,DBLP:conf/cvpr/PunnakkalCAQB21,DBLP:conf/iccv/GhoshCOTS21,DBLP:journals/corr/abs-2208-15001,DBLP:conf/eccv/GuoZWC22,DBLP:conf/eccv/DelmasWLMR22}. Natural language offers a convenient and expressive means for controlling generative models, similar to image \cite{https://doi.org/10.48550/arxiv.2204.06125} and video \cite{https://doi.org/10.48550/arxiv.2209.14792} generation. Users can specify the desired actions or poses they want the motion to exhibit, such as global transitions like running, jumping, and walking, or localized actions like throwing or kicking. They may also indicate concurrent sub-motions or sequential motions. The generated motion sequence should accurately match the prompt while appearing natural.

Determining the best-automated metric for human motion generation from natural language prompts is crucial for developing effective models. Although human judgment is considered the gold standard, comparing large sample sizes is time-consuming and expensive. Stochasticity in recent models adds to this challenge, necessitating extensive repetitions for accurate results.

Our objective is to identify the best automated metric for evaluating language-conditioned human motion generations, with "best" referring to the metric most closely correlated with human judgments. While various automated metrics have been proposed \cite{DBLP:conf/3dim/AhujaM19,DBLP:conf/iccv/GhoshCOTS21,DBLP:conf/cvpr/GuoZZ0JL022} and some works have conducted comparative human evaluations \cite{DBLP:conf/cvpr/GuoZZ0JL022,DBLP:conf/eccv/PetrovichBV22}, none have directly addressed this question. Developing appropriate automated metrics correlated with human judgments has been vital in fields such as machine translation \cite{Papineni2002BleuAM,https://doi.org/10.48550/arxiv.1904.09675}, and we believe it is essential for advancing text-to-motion methods.

To complement existing metrics, we propose novel ones that improve correlation with human judgment while being differentiable and capable of enhancing optimization when integrated into training losses. One novel metric, a multimodal BERT-like model MoBERT, offers sample level evaluation scores with significantly improved human judgment correlations. 

Multiple distinct aspects should be considered when assessing the quality of generated human motions. We evaluate human motion quality by focusing on the following:

\begin{itemize}
\item \textbf{Naturalness}: How realistic is the motion to a viewer? Unnatural motions exhibit inhuman or improbable poses or display global transitions without appropriate actions.
\item \textbf{Faithfulness}: How well does the generated motion align with the natural language prompt? Unfaithful motions will omit key components or include irrelevant ones.
\end{itemize}

Our main contributions are:
\begin{itemize}
\item A dataset of motion-text pairs with human ratings of \textit{Naturalness} and \textit{Faithfulness} for evaluating automated metrics.
\item A critical evaluation of existing text-to-motion automated metrics based on correlation with human judgments.
\item The development of novel high-performing automated metrics, including MoBERT, offering the first strongly human-correlated evaluation metric for this task. We also discuss how MoBERT addresses limitations of existing metrics, advancing future architecture comparison and development.
\end{itemize}

\footnote{Our metric evaluation code and collected human judgment dataset are included as supplemental material to this work. Our novel evaluator model, MoBERT, is available at \url{https://github.com/jvoas655/MoBERT}.}

\section{Related Works}
We review prior research on human motion generation, which includes both unconditioned and conditioned generation, and discuss the evaluation metrics used in previous studies.

\subsection{Human Motion Generation}
Early unconditioned human motion generation approaches employed statistical generative models \cite{10.1145/1477926.1477927,Mukai2005GeostatisticalMI}, while more recent models have adopted deep learning techniques. Some studies have applied Variational Autoencoder (VAE) models \cite{https://doi.org/10.48550/arxiv.1312.6114} for motion forecasting based on historical fragments \cite{https://doi.org/10.48550/arxiv.1707.04993, Aliakbarian_2020_CVPR, Ling_2020, rempe2021humor}. Others have used Generative Adversarial Networks (GAN) \cite{https://doi.org/10.48550/arxiv.1406.2661} to enhance the quality of generations \cite{https://doi.org/10.48550/arxiv.1711.09561}. Normalization Flow Networks have also been explored \cite{Henter_2020}. The majority of these methods employ joint-based frameworks, utilizing variants of the SMPL \cite{10.1145/2816795.2818013} body model, which represents the body as a kinematic tree of connected segments.

For conditioned motion generation, various types of conditioning exist. Some studies have conditioned on fixed action categories, which simplifies the task compared to natural language conditioning but limits diversity and controllability. \cite{Guo_2020} employs a recurrent conditional VAE, while \cite{https://doi.org/10.48550/arxiv.2104.05670} uses a category-conditioned VAE with Transformers \cite{https://doi.org/10.48550/arxiv.1706.03762}.

Natural language conditioning allows for fine-grained motion control, enabling temporal descriptions and specification of individual body parts. Early efforts utilized a Seq2Seq approach \cite{Lin2018GeneratingAV}. Other studies learned a joint embedding projection for both modalities \cite{DBLP:conf/3dim/AhujaM19,DBLP:conf/iccv/GhoshCOTS21} and generated motions using a decoder. Some research applied auto-regressive methods \cite{DBLP:conf/cvpr/GuoZZ0JL022}, encoding text and generating motion frames sequentially. Recent approaches, such as \cite{DBLP:conf/eccv/PetrovichBV22}, use stochastic for diverse generations. Others employed diffusion-based models \cite{DBLP:journals/corr/abs-2209-00349}\cite{DBLP:journals/corr/abs-2208-15001}\cite{https://doi.org/10.48550/arxiv.2209.14916}\cite{Wei2023UnderstandingTM}\cite{Chen2022ExecutingYC}\cite{Shafir2023HumanMD}\cite{Zhang2023ReMoDiffuseRM}\cite{Han2023AMDAM}. Recent models have taken inspiration from GPT-like LLM's through learned motion vocabularies and have competed with diffusion methods for SOTA performance \cite{zhang2023motiongpt}\cite{jiang2023motiongpt}\cite{Zhou2022UDEAU}\cite{Zhang2023T2MGPTGH}.

Related tasks have also been investigated, such as \cite{DBLP:journals/corr/abs-2008-08171} or \cite{tseng2022edge}, which conditions motion generation on music. Some models treat the task as reversible, captioning motions and generating them from language prompts \cite{DBLP:conf/eccv/GuoZWC22}. Others generate stylized character meshes to pair with the generated motions, conditioned on language prompt pairs \cite{DBLP:conf/eccv/YouwangKO22, https://doi.org/10.48550/arxiv.2205.08535}. Adjacent efforts have focused on scene or motion path-based conditioning, allowing for high-quality animation of character movements along specific paths in an environment \cite{10.11453072959.3073663}\cite{ling2020character}\cite{huang2023diffusion}.

\subsection{Metrics for Automated Evaluation of Human Motions}
Various metrics have been used to evaluate text-to-motion. \cite{DBLP:conf/3dim/AhujaM19} employed Average Position Error (APE) and pioneered the practice of dividing joints into sub-groups for different versions of APE. \cite{DBLP:conf/iccv/GhoshCOTS21} introduced Average Variance Error and also considered versions dependent on which joints (root versus all) are being used and whether global trajectories are included. \cite{DBLP:conf/eccv/PetrovichBV22} and \cite{DBLP:journals/corr/abs-2209-00349} adopted similar methods, but recent works have moved away from these metrics despite no study establishing them as poor performers.

\cite{DBLP:conf/cvpr/GuoZZ0JL022} developed a series of metrics based on their previous work for category-conditioned motion generation, advocating for Frechet Inception Distance (FID) \cite{NIPS2017_8a1d6947}, which is commonly used in image generation and measures output distribution differences between datasets. \cite{DBLP:conf/cvpr/GuoZZ0JL022} also included R Precision, a metric based on retrieval rates of samples from batches using embedded distances, metrics to evaluate diversity, as well as one measuring the distance of co-embedding in each modality. These metrics have become standard, used by \cite{DBLP:conf/eccv/GuoZWC22, https://doi.org/10.48550/arxiv.2209.14916, DBLP:journals/corr/abs-2209-00349, DBLP:journals/corr/abs-2208-15001}. These metrics rely on a text and motion co-encoder, so proving the effectiveness of the encoder is crucial for these metrics if they are to be used for judging model performance. \cite{Yuan2022PhysDiffPH} expanded these metrics to measure factors of physical motion plausibility. 

The GENEA Challenge \cite{10.1145/3397481.3450692} provides a collective assessment of co-speech motion generation methods through standardized human evaluations. It divides human judgments into \textit{Human-likeness} and \textit{Appropriateness}, corresponding to our \textit{Naturalness} and \textit{Faithfulness}. Recent findings by \cite{10.1145/3536221.3558058} indicate that current methods generate natural motions at or above rates for baseline captures but underperform in faithfulness. While not directly applicable to text-to-motion, this research provides valuable data for understanding the performance of current methods and guiding future work in the area, including novel metrics.

\section{Dataset Collection}

\subsection{Baseline Models Evaluated}

We evaluate four implementations to assess a range of motion qualities and focus on issues relevant to top-performing models: \cite{DBLP:conf/cvpr/GuoZZ0JL022}, TM2T \cite{DBLP:conf/eccv/GuoZWC22}, MotionDiffuse \cite{DBLP:journals/corr/abs-2208-15001}, and MDM \cite{https://doi.org/10.48550/arxiv.2209.14916}. These models, trained on the HumanML3D dataset \cite{DBLP:conf/cvpr/GuoZZ0JL022}, support 22 joint SMPL body models \cite{10.1145/2816795.2818013}, enabling consistent animation methods for human ratings. We also include reference motions from HumanML3D as a baseline for non-reference evaluation metrics. 

\subsection{Motion Prompt Sample Collection}

We sourced motion prompts from the HumanML3D test set. To ensure diverse and representative prompts, we encoded them using the RoBERTa language model's CLS outputs \cite{liu2019roberta}. The embeddings were projected onto a low-dimensional space and we randomly sample from the resulting dataset's distribution, taking the nearest unsampled entry, to obtain 400 unique sample prompts. 

These prompts generated a dataset of 2000 motions, with 400 motions for each of the five baseline models (including HumanML3D). For models generating fixed-length motions, we used a length of 120 motion frames. All models were generated at the 20 Hz frequency used in HumanML3D.

\subsection{Motion Visualization}

Recent studies \cite{DBLP:conf/cvpr/GuoZZ0JL022, DBLP:conf/eccv/PetrovichBV22} utilized stick figure renderings for evaluation, but this approach has limitations. Evaluating \textit{Naturalness} using stick figures can be challenging, as they are not relatable to raters. Moreover, they often lacked realistic environments, such as walls, floors, lighting, and textures.

To address these limitations, we created high-quality renders using Blender \cite{blender}, focusing on environmental details and camera movements for natural motion perception (Figure \ref{fig:motion-exps}).

\subsection{Human Quality Ratings Collection}

We collected human quality ratings using Amazon Mechanical Turk and a custom UI. To ensure quality, we implemented qualification requirements, in-tool checks, and post-quality criteria. We hand-picked 25 motion-text pairs from the 2000 motion samples we generated and used them as gold test questions\footnote{Gold test questions ground truth labels were judged by the Authors. Motions for which the ratings were deemed to be overly subjective were not included in the gold test set.}. The remaining annotations were divided into 20-pair batches, each containing five randomly placed gold test samples. We collected three ratings per sample and discarded batches that failed qualification checks.

Ratings were presented as natural language descriptions corresponding to Likert Scale ratings (0 to 4). Annotators had access to a tooltip with detailed descriptions for each rating level during the task, all shown in Figures \ref{fig:mturk_inst}, \ref{fig:mturk_view}, and \ref{fig:mturk_rate} of the supplement.

Ratings were rejected if more than two of the five test questions deviated by more than one from the "correct" answer. This allowed for subjectivity, missed details, and slight rating scale understanding differences. Significant deviations in rating scale understanding or guessing would pass a single question occasionally, but over the ten independent ratings would be detected with a high likelihood. In-tool quality checks required watching the entire video before progressing and capped the rate of progression to 12 seconds per sample. These measures aimed to prevent rushing and encourage thoughtfulness. Qualification requirements included residing in the U.S., completing over 1000 hits, and a minimum $98\%$ acceptance rate. Quality checks were disclosed in the task instructions. We paid \$1.25 per HIT, equating to at least \$12 per hour.

\begin{figure}[H]
\centering
\includegraphics[width=\linewidth]{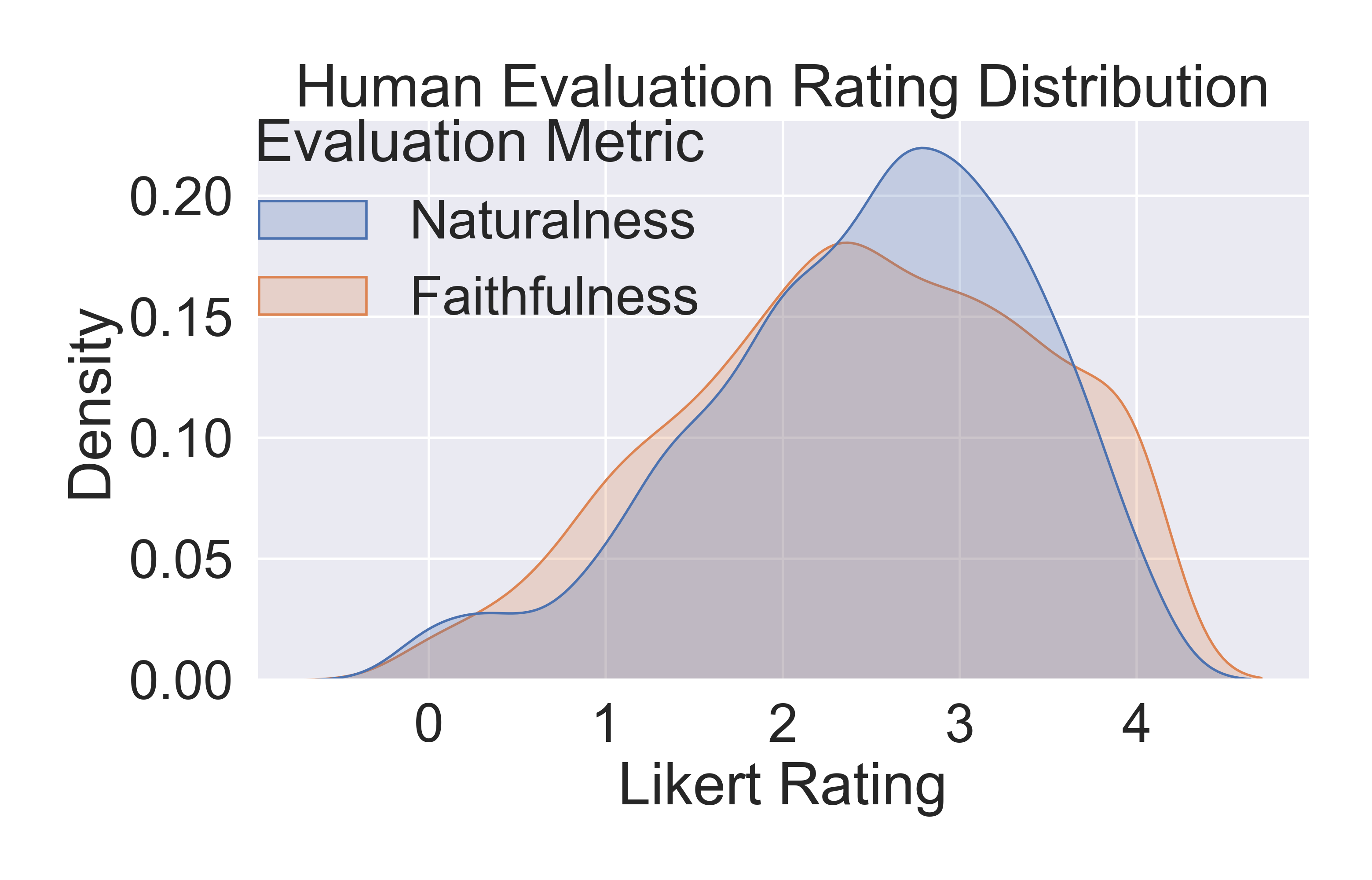}
\caption{Human judgment distribution for all samples. Averages from three annotations are shown with a KDE smoothing filter (bandwidth 0.85) applied. Pearson's correlation between metrics is found to be 0.63 at the sample level.}
\label{fig:rating-dist}
\end{figure}

We removed samples with less than three ratings for all models, resulting in 1400 rated motion-text pairs (280 distinct prompts for each baseline model). Averaging the three ratings provided final \textit{Naturalness} and \textit{Faithfulness} values. We show in Figure \ref{fig:rating-dist} the dataset's distribution to be generally normal, while Table \ref{fig:rating-iaa} shows high inter-annotator agreement (Krippendorff's Alpha) was obtained.

\section{Evaluated Metrics}

We evaluate most automated metrics from recent works as well as  new ones. We assess each metric's correlation with samples on both individual and model levels, whenever possible. Sample level correlations are computed on individual sample scores across baselines, reflecting the metric's capability to evaluate individual generations. Model-level correlations are determined using the mean metric score for all samples generated by a specific baseline model, which are then correlated with the mean human rating for the corresponding samples. This assesses how well the metric can judge model performance ranking. These levels can be distinct since metrics with outlier failures may negatively impact sample-level evaluation but have reduced effects when averaged over many samples. 

To calculate FID, R-Precision, and Multimodal Distance the motion features must be projected into an embedding space using an encoder. The encoder used was developed by \cite{DBLP:conf/cvpr/GuoZZ0JL022} and is standard for these metrics. 

\subsection{Existing Metrics}

\subsubsection{Coordinate Error (CE) Metrics}

Average Error (AE), also known as Average Position Error (APE) when applied to joint positions \cite{DBLP:conf/3dim/AhujaM19}, and Average Variance Error (AVE) \cite{DBLP:conf/iccv/GhoshCOTS21} are reference-based metrics employed in early works but have become less common recently. They calculate the mean L2 errors between reference and generated values, either absolute or as variance across frames, for each joint in the motion. We refer to these as coordinate error (CE) metrics, defined as:

\begin{equation}
    AE = \frac{1}{JT} \sum_{j \in J}\sum_{t \in T} \|X_{t}[j] - \hat{X}_{t}[j]\|_{2}
\end{equation}
\begin{equation}
    \sigma[j] = \frac{1}{T - 1} \sum_{t \in T} (X_{t}[j] - \bar{X}_{t}[j])^{2}
\end{equation}
\begin{equation}
    AVE = \frac{1}{J} \sum_{j \in J} \|\sigma[j] - \hat{\sigma}_{t}[j]\|_{2}
\end{equation}
Where $j$ represents a joint from all 22 joints $J$, and $t$ denotes a motion frame from the motion sequence $T$. We matched frame lengths for reference and generated motions by clipping the longer one.

We investigate CE metrics on positional values and their variations on positional derivatives, such as velocity and acceleration, calculated using frame-wise differences. Additionally, we evaluate these metrics on combinations of position and its derivatives. Similar to \cite{DBLP:conf/iccv/GhoshCOTS21}, we consider three joint groupings for CE metrics: root only, all joints excluding the root (Joint), and all joints (Pose). Prior works \cite{DBLP:conf/iccv/GhoshCOTS21, DBLP:conf/3dim/AhujaM19} suggested that AE on the root joint best aligns with quality.

We hypothesize that this effect might stem from scaling issues when the root translations are included in combined calculations with other joints, causing their errors to dominate the metric. To test this, we explore potential root joint scaling factors, altering their transitions contribution to the metric's final score for the mean. We also examined the impact of scaling factors on each component when calculating combined position-velocity (PV) or position-velocity-acceleration (PVA) CE. Component-based scaling acts as a weighted average, with scaling factors increasing or decreasing the component errors, while root scaling adjusts the effects of root translation on all joint positions.

\subsubsection{Fréchet Inception Distance (FID)}
The Fréchet Inception Distance (FID) \cite{NIPS2017_8a1d6947} is a widely used metric for generative tasks, which measures the alignment between two distributions. To compute FID, one must first obtain the mean and variance of each distribution from a large sample size. In generative tasks, these typically correspond to the reference samples (a valid distribution) and the generative model samples. A lower FID indicates better alignment between the generative and reference distributions. FID is calculated as follows for distributions $D_{1}$ and $D_{2}$:

\begin{equation}
FID(D_{1}, D_{2}) = |\mu_{1} - \mu_{2}| + tr(\Sigma_{1} + \Sigma_{2} - 2 (\Sigma_{1}\Sigma_{2})^{\frac{1}{2}})
\end{equation}

As FID is only accurate with large sample sizes, we report correlations for FID at the model level only and do not report correlation scores for individual samples.

\subsubsection{R-Precision}

R-Precision is a distance-based metric that measures the rate of correct motion-prompt pair matchings from a batch of random samples. Both motions and prompts are projected into a co-embedding space, and Euclidean Distance calculations are used to rank pair alignments. Scores of one are received if the correct matching is made within a rank threshold (Retrieval Allowance), and zero otherwise. Averaged over numerous samples, this provides a precision of retrieval metric.

Higher Retrieval Allowance thresholds yield higher R-Precision scores, as they are more forgiving of imperfect embedding spaces and account for multiple motions described by the same prompt randomly being included in the batch. R-Precision scores for thresholds of 1-3 are commonly reported. We analyze the correlation for R-Precision scores with thresholds of 1-20 and hold the batch size to 32, following common practice \cite{DBLP:conf/cvpr/GuoZZ0JL022}.

\subsubsection{Multimodal Distance}
This metric measures the distance between the generated motion embedding and the co-embedding of the prompt used for generation. When the two encoders are well-aligned in the embedding space, low scores suggest motions closely matching the prompt, while high scores indicate significant deviations in features \cite{DBLP:conf/cvpr/GuoZZ0JL022}.

\begin{figure}[t]
\centering
\includegraphics[width=\linewidth]{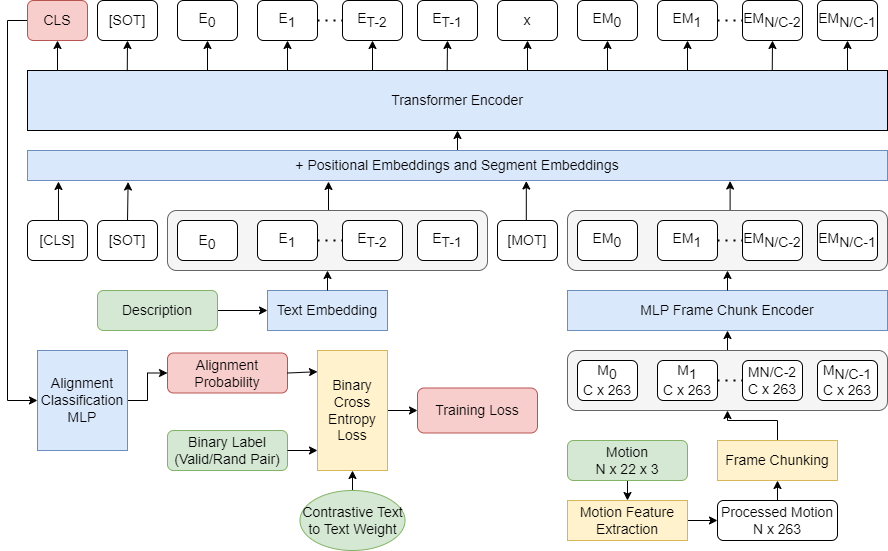}
\caption{Our MoBERT architecture and process flow. Green items represent inputs, white items indicate intermediate steps, red items denote output/losses, and blue items contain learned model parameters.}
\label{fig:model_outline}
\end{figure}

\subsection{MoBERT: Multimodal Transformer Encoder Evaluator}

Our novel evaluation method, MoBERT, is inspired by past learned metrics such as CLIPScore \cite{hessel2022clipscore}, that score the alignment between a multimodal pair. However, MoBERT distinguishes itself by its ability to evaluate both modalities using a shared Transformer Encoder \cite{https://doi.org/10.48550/arxiv.1706.03762} through a multimodal sequence embedding. This approach, as shown in Figure \ref{fig:model_outline}, employs the attention mechanism of the Transformer to capture detailed relationships between the motion chunks and textual tokens.

Compared to CLIPScore, which uses separate encoders for each modality and combines the two modalities using cosine similarity, MoBERT's single Encoder approach allows for a richer understanding of the data. The Transformer Encoder's attention mechanism can learn to consider features across both modalities simultaneously, potentially capturing nuanced relationships between them that might be missed in a separate encoding scheme. In particular, this methodology allows MoBERT to consider the shared temporal aspects of motions and text prior to being collapsed to a single vector representation. This approach allows for more accurate prediction of correct and incorrect text pairings, allowing MoBERT to potentially outperform methods following CLIPScore's approach. 

\subsubsection{Encoding Motion Information}

To better contextualize motion in our model, we preprocess our $N \times 22 \times 3$ motions into an $N \times 263$ representation following the approach in \cite{DBLP:conf/cvpr/GuoZZ0JL022}. This involves extracting motion transformations, such as root joint global transitions and rotations, to handle shifts in reference frames, as well as the linear velocities of each joint frame-to-frame and foot contact thresholding for a binary signal of foot-ground contact.

To utilize frame-to-frame motion information and mitigate redundancy in the motion domain, we downsample encodings by chunking consecutive frames into frame chunks before converting them into embeddings. Our dataset motions span up to 200 frames, processed at 20 Hz. We group these into 14-frame chunks, as 0.7 seconds of motion information offers adequate encoding and information differentiation. To account for the simplicity of our chunking algorithm, we apply an overlap factor of 4 frames, duplicating overlapped frames in consecutive motion chunks.

\subsubsection{Multimodal Tokenization Process}
For encoding text, we utilize a BPE \cite{Gage1994ANA} vocabulary and learned embeddings. We generate sequence embeddings from the textual and motion processes and merge them into a single sequence (Figure \ref{fig:model_outline}). We incorporate special tokens for CLS, start indicators, and padding embeddings. With short one or two-sentence descriptions and motions limited to a chunk length of 20, we train using a max context size of 64. Learned segment and positional tokens are added to inputs.

\subsubsection{Training Process}
We used the HumanML3D dataset as the basis for our model's training. The model is trained through the task of \textbf{Alignment prediction} using Binary Cross Entropy loss. This task involves predicting a binary label that indicates whether a given motion corresponds to a specific textual description. For each motion-text pair in our training dataset, we randomly selected a contrastive textual description to serve as a negative label example. We then evaluate both valid and contrastive pairings with the model, resulting in alignment probability judgments. We employed a compact MLP model over the CLS output embeddings, terminating with sigmoid activation, to obtain an output alignment probability. Binary Cross Entropy loss is used to encourage the model to predict alignment labels for valid pairings and anti-alignment labels for incorrect pairings, as demonstrated in Equations \ref{eq:bce} and \ref{eq:l1}.

\begin{equation}
\begin{aligned}
H(q) = -\frac{1}{N} \sum_{i=1}^{N} y_{q}(i) \cdot log(p(y_{q}(i), q))\\ + (1 - y_{q}(i)) \cdot log(1 - p(y_{q}(i), q))
\end{aligned}
\label{eq:bce}
\end{equation}
\begin{equation}
\mathscr{L}1 = H(V) + H(R)
\label{eq:l1}
\end{equation}
With $N$ being all motions in a batch, $y_{q}(i)$ is the correct binary label for sample $i$ given text grouping $q$ (valid or contrastive), and $p$ being the predicted binary label. $V$ is the set of valid textual descriptions and $R$ is the set of random contrastive descriptions. We found that this process could still present a difficult optimization landscape, and would often choose to predict all one label to minimize loss on one pairing despite increased losses on the other. To promote balancing each label's prediction, we achieved better results with the L2 balanced loss shown in Equation \ref{eq:l2}.

\begin{equation}
\mathscr{L}_2 = \sqrt{H(V)^2 + H(R)^2}
\label{eq:l2}
\end{equation}

Additional tasks, in a multi-task learning framework, were trialed but did not improve performance and were not included in the version of MoBERT's we report in this work. 

\paragraph{Improving Contrastive Examples}
The HumanML3D dataset provides low diversity of descriptions, with many being very similar. Further, motions can be described in multiple ways, both of which make random contrastive textual samples provide low-quality guidance. To address this, we used Sentence Transformer similarity scores to weight contrastive training examples and adjust our loss functions accordingly. Inverse similarity scores were applied as weights to the loss function, down-weighting similar descriptions to reduce label confusion. We employed the top-performing Huggingface "all-mpnet-base-v2" implementation. The contrastive loss was rescaled by the weights to maintain a consistent magnitude with the valid loss. The final loss function is shown in Equation \ref{eq:lf}, where alpha represents the similarity scores produced by the Sentence Transformer model score, confined to $[0, 1]$.

\begin{equation}
\mathscr{L}_f = \sqrt{H(V)^2 + \Biggl(\frac{(1 - \alpha) H(R)}{\sum_{i}^{N} (1 - \alpha_{i})}\Biggr)^2}
\label{eq:lf}
\end{equation}

\paragraph{Model Evaluation Process}
We assess the correlation of our baseline models' raw Alignment Probability scores from our training process. Since this data lacks human rating guidance, we also test our model's performance when trained on a small set of human judgment data. We do this by discarding the output layers of our model, using an aggregation of output embeddings, and fitting a lightweight SVR or Linear Regression layer to predict human judgments. The best performance is achieved using a RBF Kernel SVR, with a Ridge regressor being the best fully differentiable. Sklearn's Python package is used for regression training and hyperparameters are reported in the supplemental materials section.

To avoid overfitting to the small human judgment dataset, we apply ten-fold cross-validation, fitting regressors on 90\% of the dataset's samples to predict the remaining portion. These cross-validated predictions are collected, reordered, and Pearson's correlation is calculated against the human judgment ratings.
\begin{table*}
\begin{center}
\begin{tabular}{|| l r | c | c | c | c ||} 
 \hline
  & & \multicolumn{2}{| c |}{Model Level} & \multicolumn{2}{| c ||}{Sample Level} \\
  \cline{3-6}
  \multicolumn{2}{|| c |}{Metric} & Faithfulness & Naturalness & Faithfulness & Naturalness \\[0.5ex]
  \hline\hline
  Root AVE & $\downarrow$ & -0.926 & -0.908 & -0.013 & 0.007 \\
 Root AE & $\downarrow$ & -0.715 & -0.743 & -0.033 & 0.037 \\
 Joint AVE & $\downarrow$ & -0.260 & -0.344 & -0.178 & -0.185 \\
 Joint AE & $\downarrow$ & -0.120 & -0.227 & -0.208 & -0.245 \\
 \hline
 Multimodal Distance & $\downarrow$ & -0.212 & -0.299 & 0.025 & 0.014 \\
 R-Precision & $\uparrow$ & 0.816 & 0.756 & 0.036 & 0.042 \\
 FID & $\downarrow$ & -0.714 & -0.269 & - & - \\
 \hline
 MoBERT Score (Alignment Probability) & $\uparrow$ & \textbf{0.991} & 0.841 & 0.488 & 0.324 \\
 MoBERT Score (SVR Regression)* & $\uparrow$ & 0.962 & \textbf{0.986} & \textbf{0.624} & \textbf{0.528} \\
 MoBERT Score (Linear Regression)* & $\uparrow$ & 0.951 & 0.975 & 0.608 & 0.515 \\
\hline
\end{tabular}
\end{center}
\caption{Pearson correlations with human judgments calculated for several existing metrics and our MoBERT model. The best-performing metric in each category is bolded. Models with (*) were judged through 10-fold cross-validation. R-Precision scores reported used the best settings identified (2 for sample level, 5 for model level). Arrows next to metrics indicate whether negative ($\downarrow$) or positive ($\uparrow$) correlation is expected. }
\label{fig:results-corrs}
\end{table*}

\section{Results Analysis}
This section highlights the key findings from our evaluation. We employed Pearson's Correlation Coefficient \cite{Sedgwicke4483} to correlate metrics with human judgments, measuring the linear relationship between metrics as most of our data is interval rather than ordinal. We present model and sample level correlations between \textit{Faithfulness} and \textit{Naturalness} in Table \ref{fig:faith-nat-corr}. 

All values are uncorrected, and negative correlations are expected for certain metrics (FID or CE) since our human judgment ratings suggest better outcomes with opposing directions. Weak P-values are observed for many reported correlations, which is anticipated as they were calculated (for model level results) based on only five samples. Our strongly-performing metrics achieved P-Values near 0.05 at the model level, while our best-performing sample-level metrics (Pearson's of 0.2 or above) had near zero P-Values. 

\subsection{Coordinate Error Metrics Results}

\begin{figure}[H]
\centering
\includegraphics[width=\linewidth]{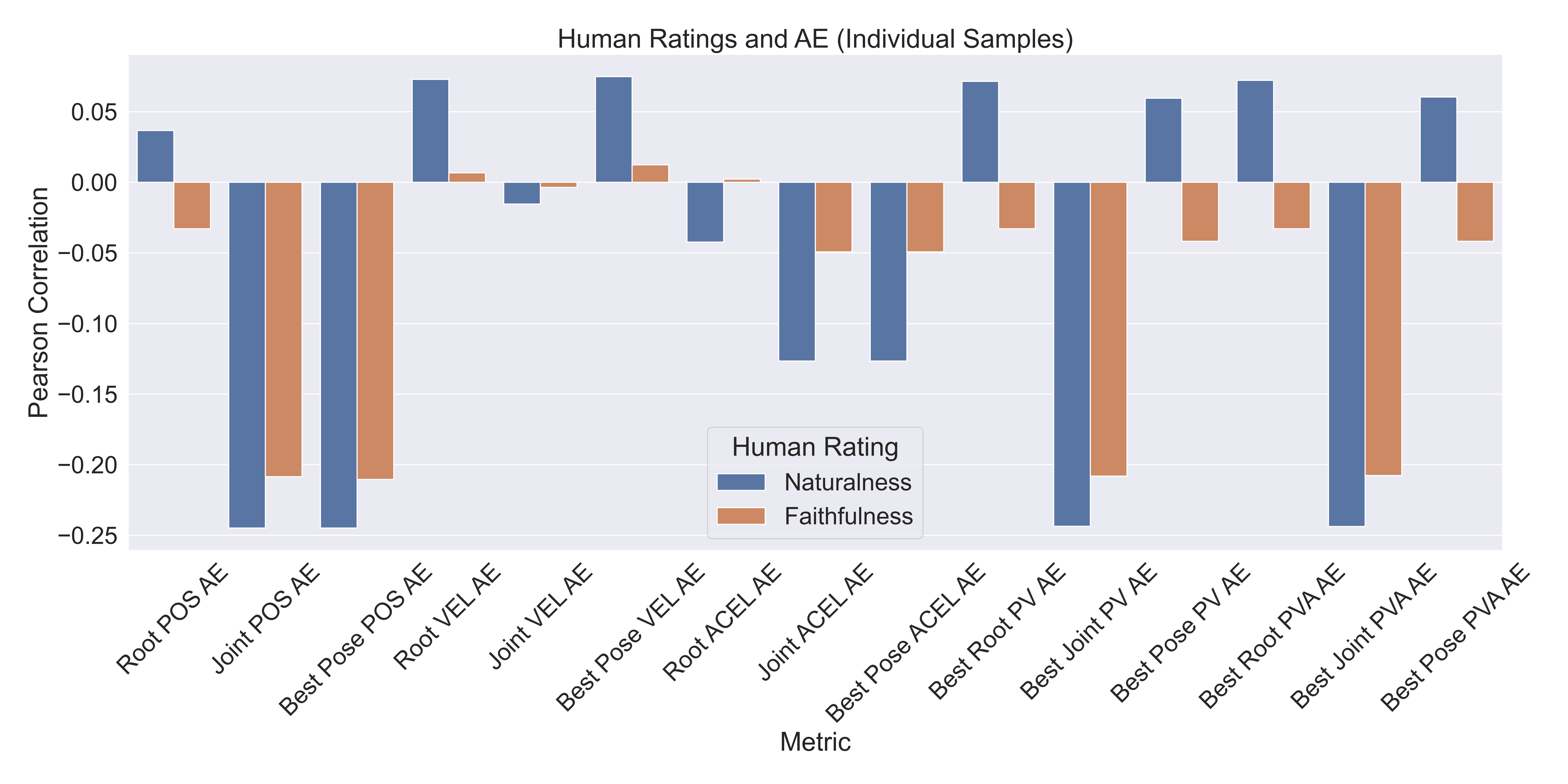}
\includegraphics[width=\linewidth]{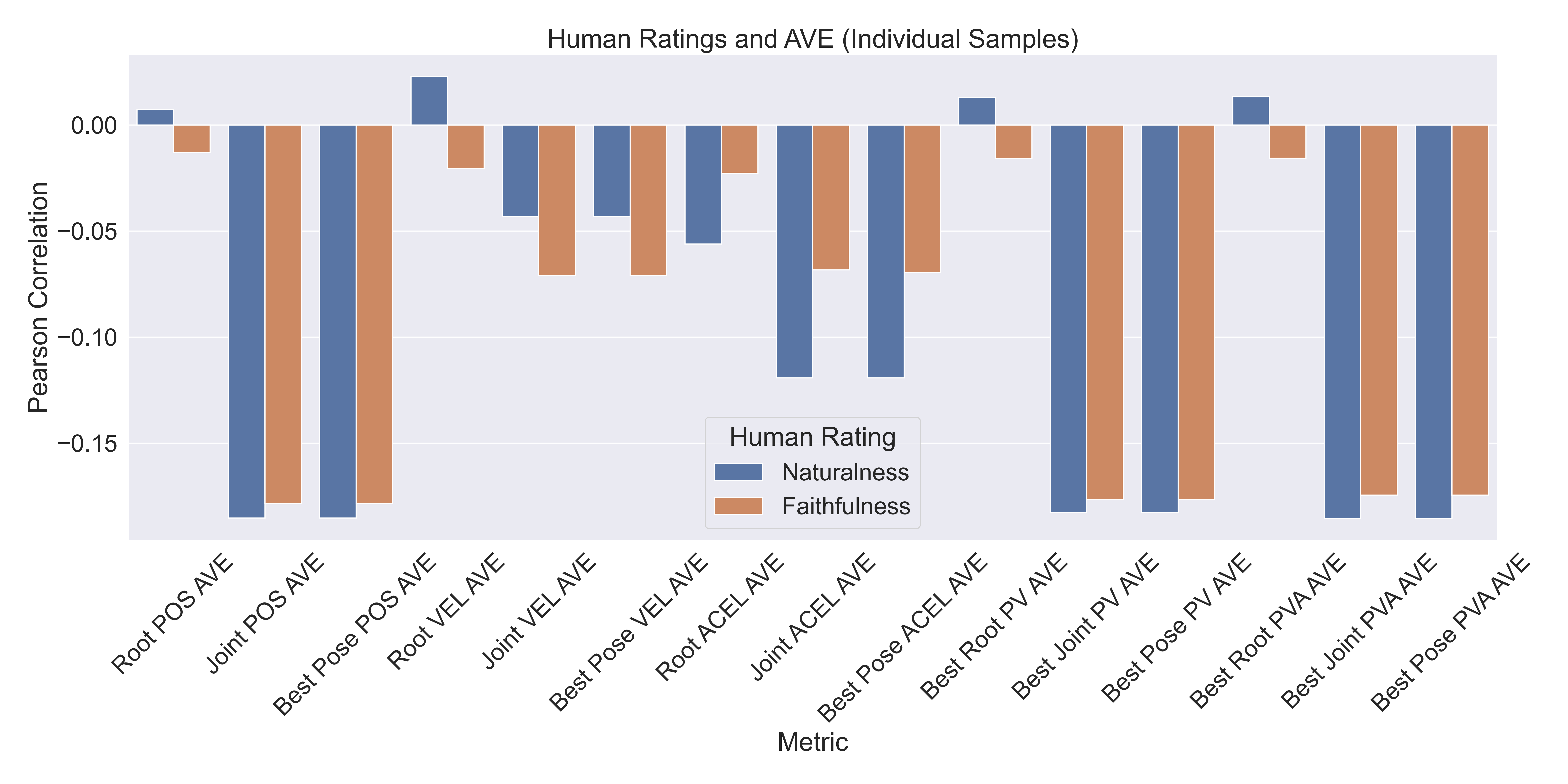}
\caption{Sample level correlations of CE metrics with human judgments. "Best" denotes versions using highest performing settings for scaling root joints or components. Greater magnitude indicates better performance. }
\label{fig:by-ind-ae-ave}
\end{figure}

\begin{figure}[H]
\centering
\includegraphics[width=\linewidth]{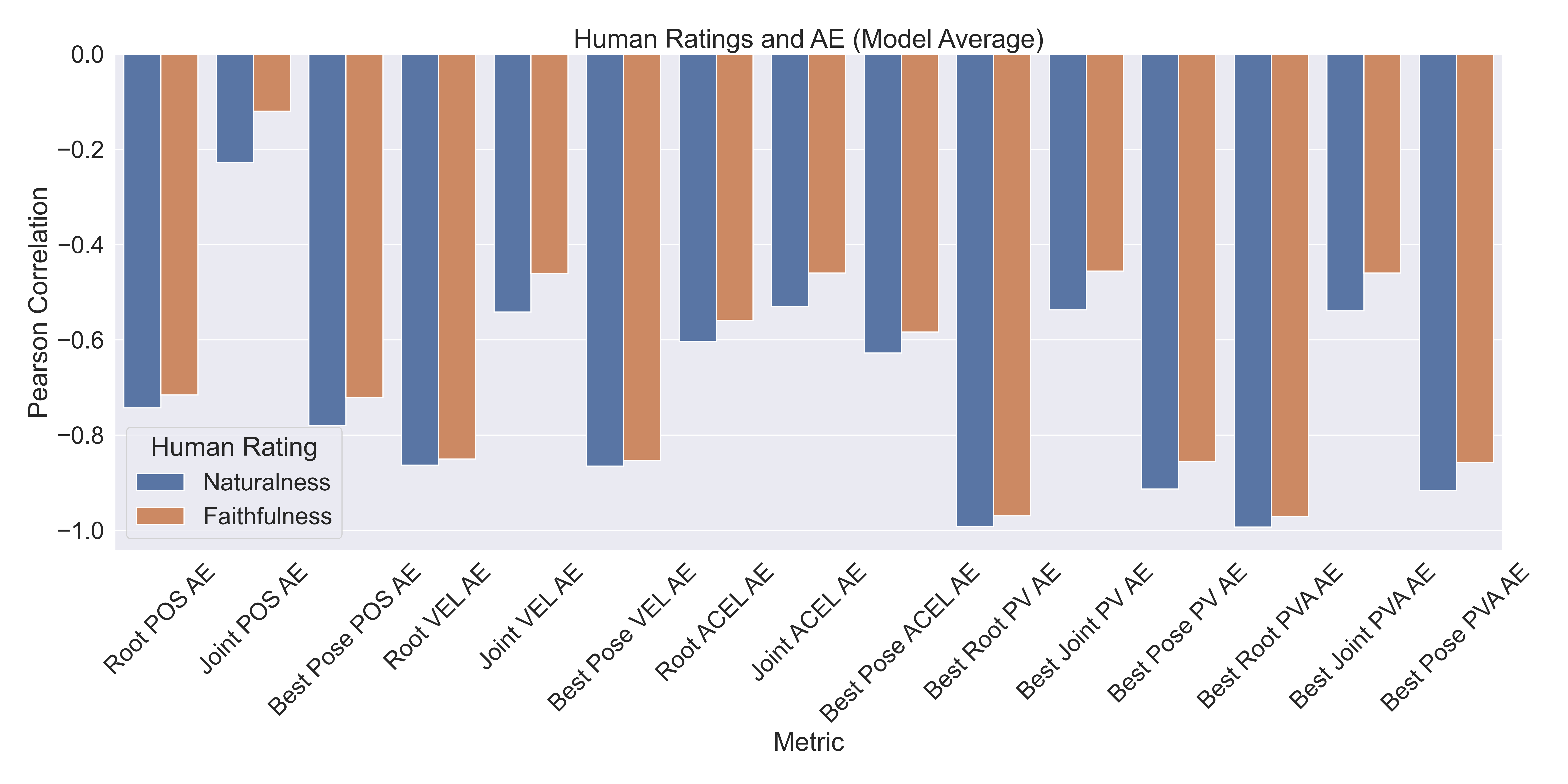}
\includegraphics[width=\linewidth]{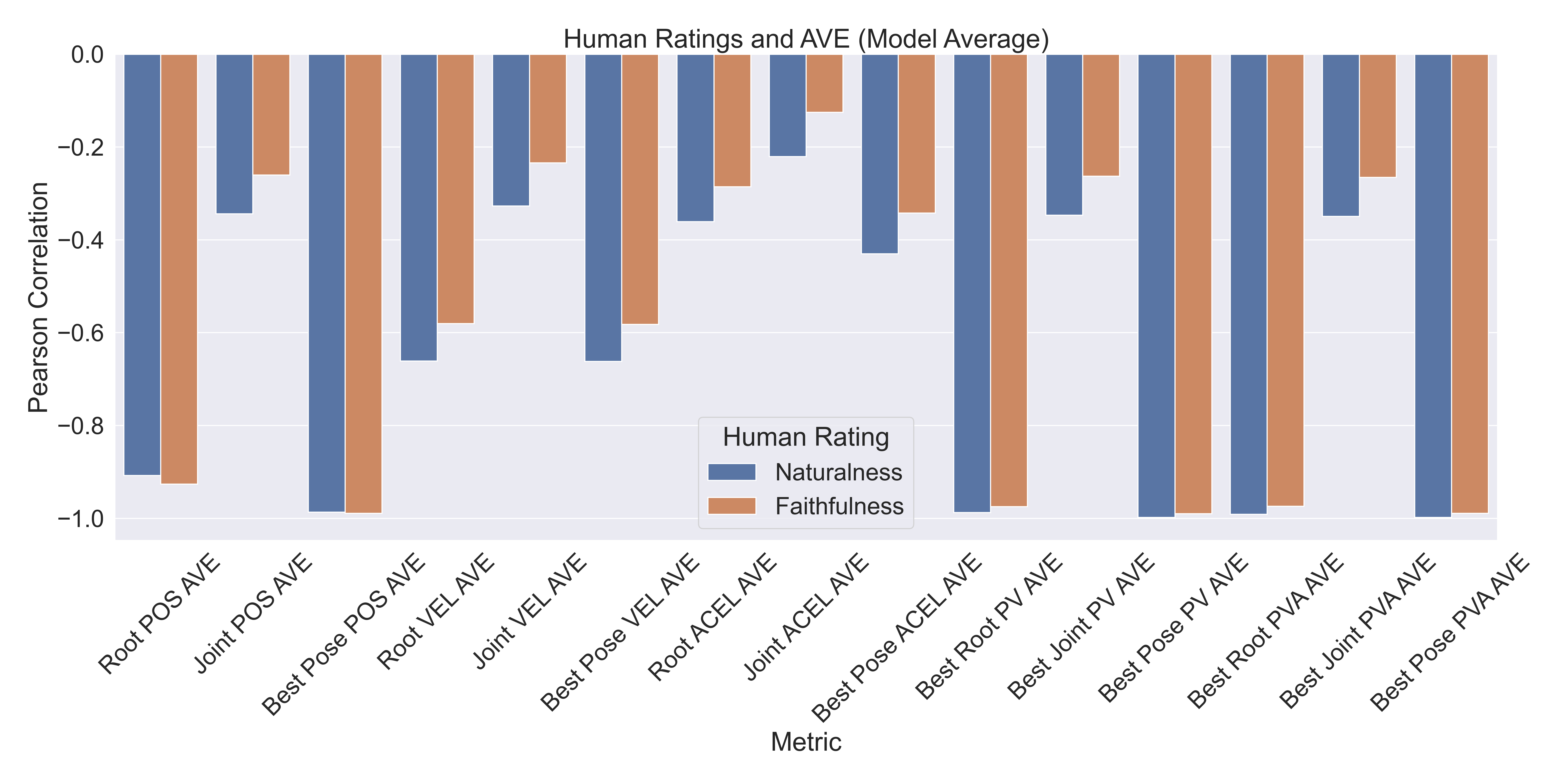}
\caption{Model level correlations of CE metrics with human judgments. "Best" metrics use the highest performing settings for root joint or component scaling. Greater magnitude indicates better performance. }
\label{fig:by-model-ae-ave}
\end{figure}

The primary CE-metric results are presented in Table \ref{fig:results-corrs} with further details in Figures \ref{fig:by-ind-ae-ave} and \ref{fig:by-model-ae-ave}. Despite relying on only a single reference, CE metrics show weak but significant correlations with human judgments for both \textit{Faithfulness} and \textit{Naturalness} at the sample level. Performance largely depends on non-Root transitions, with Joint POS AE and Joint POS AVE outperforming pure Root-based metrics. Root scaling does not surpass Joint metrics, and our derivative-based methods do worse than positional ones. Combining components only achieves results comparable to Joint POS-based metrics. Notably, AE performs better than AVE at the sample level with a significant margin (0.1 Pearson's).

At the model level, CE-based metrics strongly correlate with human judgments. Root-only traditional AE metrics achieve nearly 0.75 Pearson's, while Root AVE metrics surpass AE with approximately 0.91 Pearson's. Interestingly, Joint versions are unreliable on their own at the model level, suggesting that the main components of model evaluation can be derived from Root transitions alone. This supports similar claims by \cite{DBLP:conf/iccv/GhoshCOTS21}. Root scaling enhances both metrics, with AVE nearing perfect correlation. Utilizing velocity derivatives benefits AE at the model level, and combining positions, velocity, and/or acceleration for both AVE and AE yields versions with greater than 0.99 Pearson's (Figure \ref{fig:by-model-ae-ave}).

\subsubsection{Root Scaling Exploration}

\begin{figure}[H]
\centering
\includegraphics[width=0.92\linewidth]{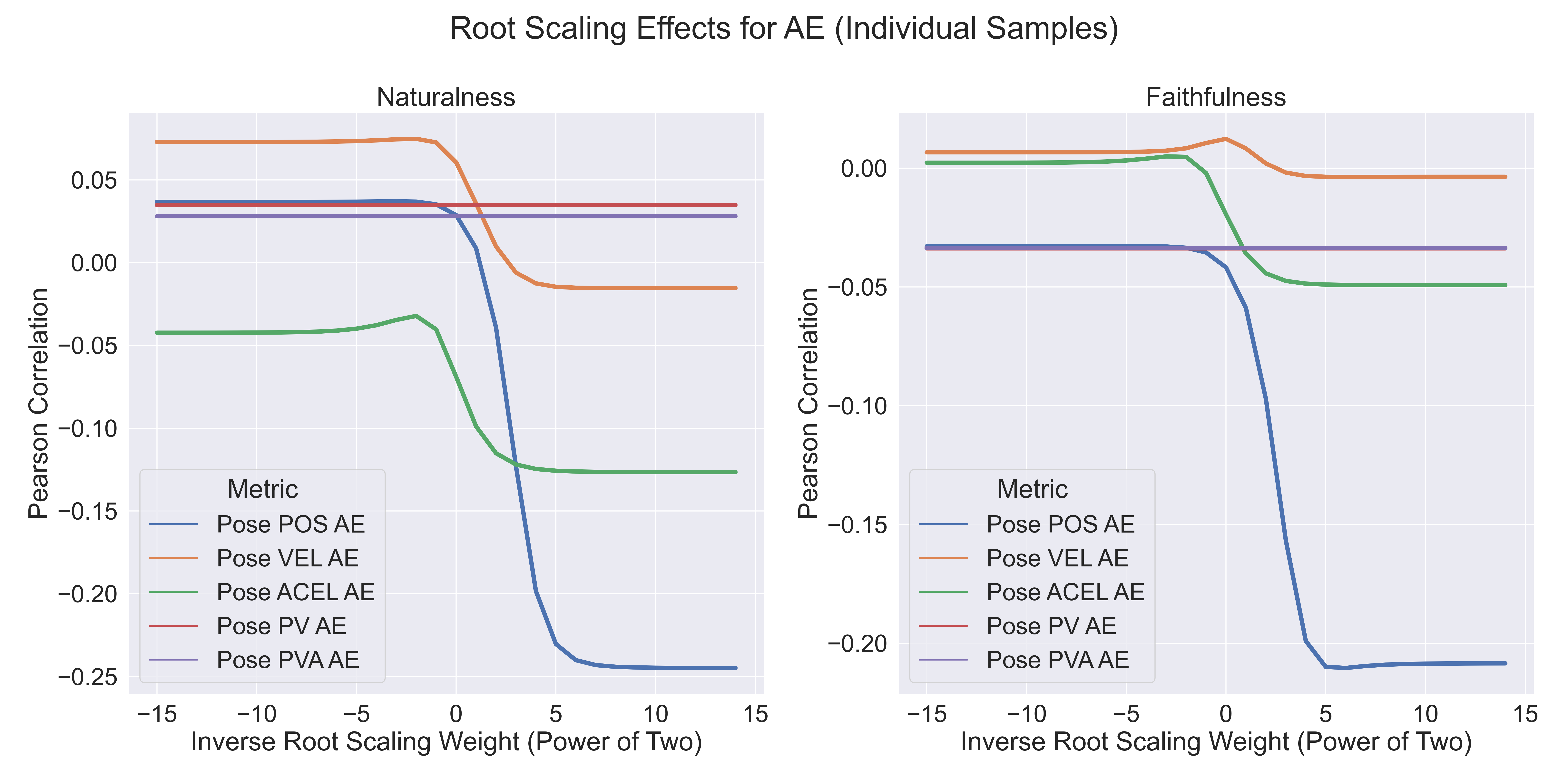}
\includegraphics[width=0.92\linewidth]{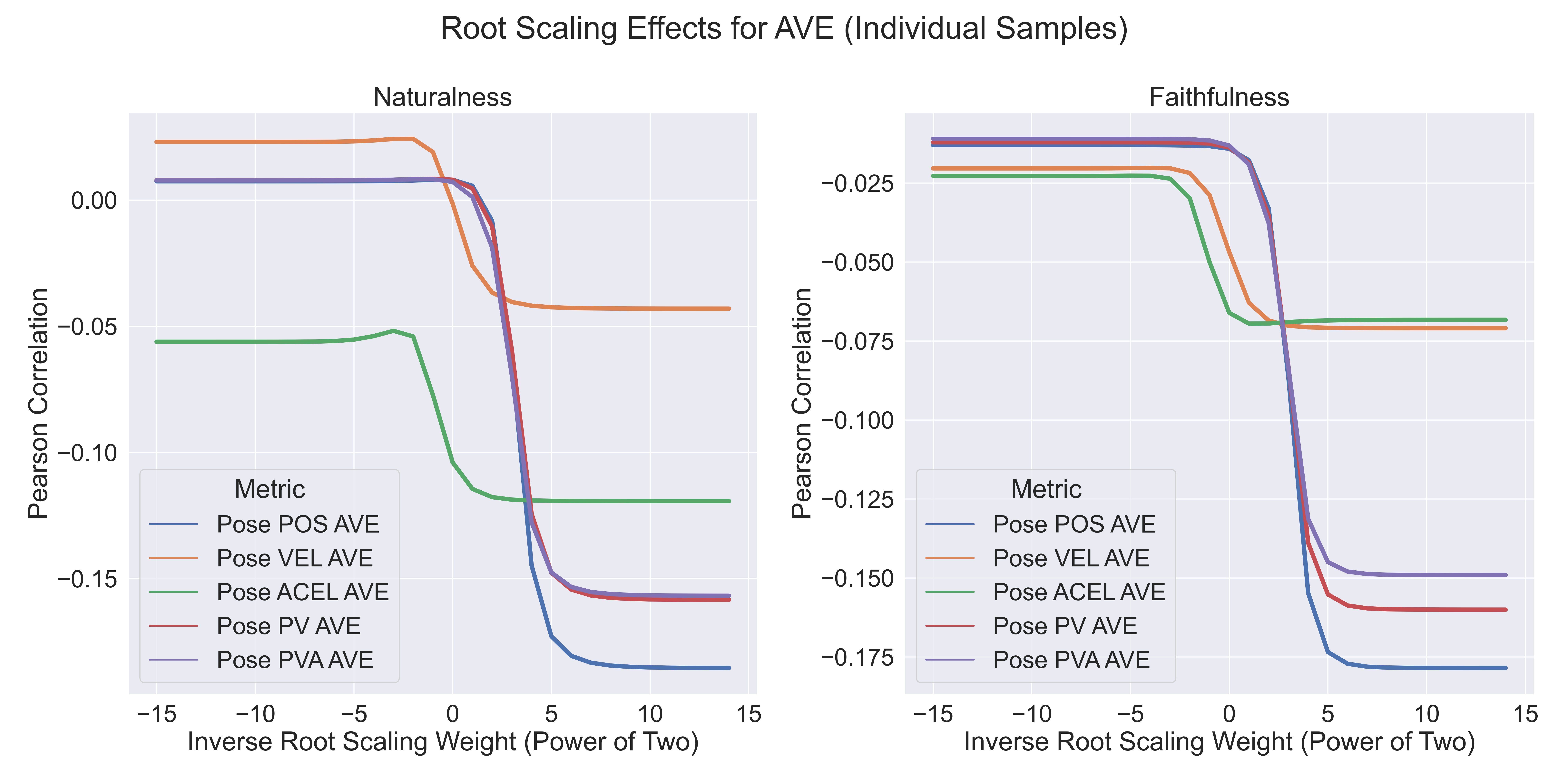}
\caption{Sample level examination of root joint scaling effects on CE using all joints. Greater magnitude indicates better performance. }
\label{fig:by-ind-root-scale}
\end{figure}

We provide visualizations with scaling factors in Figures \ref{fig:by-ind-root-scale} and \ref{fig:by-model-root-scale} to investigate the effects of root scaling on Pose CE metrics. Consistent with previous observations, model-level correlations improve (i.e., become more negatively correlated) when additional weight is placed on Root transitions. PV and PVA AE are the only versions that do not exhibit this trend. Alternatively, overemphasizing Root transitions significantly degrades performance at the sample level.

\subsection{FID, R-Precision, and Multimodal Distance Results}

\begin{figure}[H]
\centering
\includegraphics[width=\linewidth]{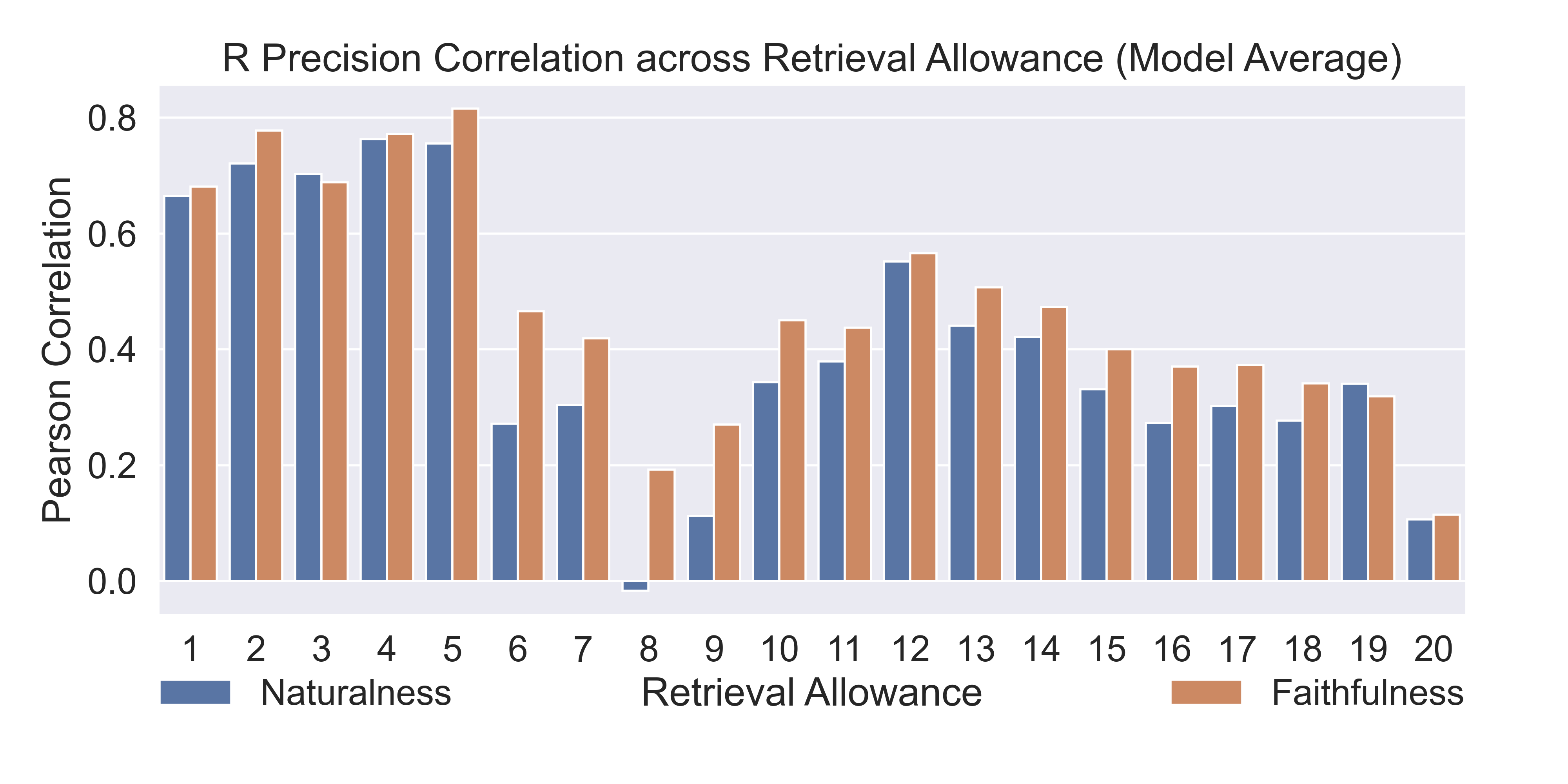}
\caption{Model level R-Precision correlations with human judgments. Retrieval Allowance indicates the number of top samples (out of a batch size of 32) considered successful if the true match is found.}
\label{fig:by-model-rp}
\end{figure}

Results for FID, R-Precision, and Multimodal Distance are also shown in Table \ref{fig:results-corrs}, with additional detail for R-Precision across various Retrieval Thresholds in Figure \ref{fig:by-model-rp}. We examine FID only at the model level as it requires distributional statistics over multiple samples, preventing sample-level calculation. We present results for R-Precision at the sample level, but R-Precision provides only binary values at this level and so it is poorly suited for sample-level comparisons with Likert ratings unless averaged over multiple samples. Multimodal Distance scored near zero at the sample level so none of these metrics provide sample-level alternatives to CE metrics.

Regarding model-level results, FID achieves acceptable results for \textit{Faithfulness} with 0.71 Pearson's but significantly underperforms for \textit{Naturalness}. Given the weak correlation with \textit{Naturalness} and model-level-only comparison, P-Values are notably weak. While these results are poor, it is possible our samples may provide an unfavorable setting for FID, or may improve with more samples.  

R-Precision demonstrates substantial correlations for both human quality judgments, approaching 0.8 Pearson's with standard settings. Our results suggest current Retrieval Thresholds are suboptimally set, with thresholds of 4 and 5 being marginally better, and then declining at higher values. Since R-Precision and FID share an embedding space, strong R-Precision results may indicate that FID's poor performance is not due to sample selection. Multimodal Distance is only weakly correlated with human quality judgments.

The results indicate that R-Precision, and possibly FID, are suitably correlated with human judgments. However, these metrics are less correlated than the CE metrics they replaced, and they preclude single-sample analysis, relying on many samples. Even if these metrics improved with larger sample sizes, an uncertain possibility, they would require substantial enhancements to match even traditional CE metrics such as Root POS AVE.

\subsection{MoBERT}

Results for our novel learned metric are shown in Table \ref{fig:results-corrs}, highlighting its performance against the best alternative metrics at the sample and model level. We observe that MoBERT substantially outperforms the best alternatives at both levels. The alignment probability outputs, without human judgment supervision, achieve a sample-level correlation of 0.488 for \textit{Faithfulness}, up from a previous best of 0.208. As expected, the correlation with \textit{Naturalness} is significantly weaker but still surpasses all other sample level correlations demonstrated by the baselines. Similarly strong results are observed for model-level performance.

Using a learned regression model over the output features further improves the results, highlighting the benefits of training on a small amount (approximately 1260 samples) of human-judgment. Our sample level correlations for \textit{Faithfulness} and \textit{Naturalness} increase to 0.624 and 0.528, respectively, reaching the strongly-correlated range for \textit{Faithfulness} when using the SVR regression layer. Moreover, our model achieves near-perfect model-level correlations, verifying that its ability to signify improved model performance is highly reliable. We run additional experiments exploring MoBERTs ability to act as a text-free Naturalness evaluator in the supplemental materials.

\subsection{Discussion and Future Work}

Our findings underscore CE metrics as the most reliable baseline metric, demonstrating strong model-level performance supported by sample-level results. With the application of root/component scaling, CE metrics reached near-perfect model-level correlations, highlighting their significance when compared with newer metrics that showed weaker performance in our study.

Although R-Precision and FID demonstrate some correlation with human judgments, their relative significance should be evaluated in context. R-Precision reveals a solid correlation, yet fall short compared with CE metrics and should be considered supplemental. FID, while showing acceptable correlations with \textit{Faithfulness} and some correlation with \textit{Naturalness}, should be used with caution in consideration of its potential to improve with more samples, but not prioritized over more consistent metrics. We recommend against the use of Multimodal Distance due to its consistently weak correlations.

MoBERT significantly outperforms all competitors, presenting the first metric with robust model-level and sample-level performance. This metric also avoids reliance on any reference motions for evaluation, making it usable in more situations and alleviating concerns about the one-to-many nature of this task. Additionally, it is fully differentiable and could be used as a training objective for generative models in order to further enhance performance. 

\begin{figure}[H]
\centering
\includegraphics[width=\linewidth]{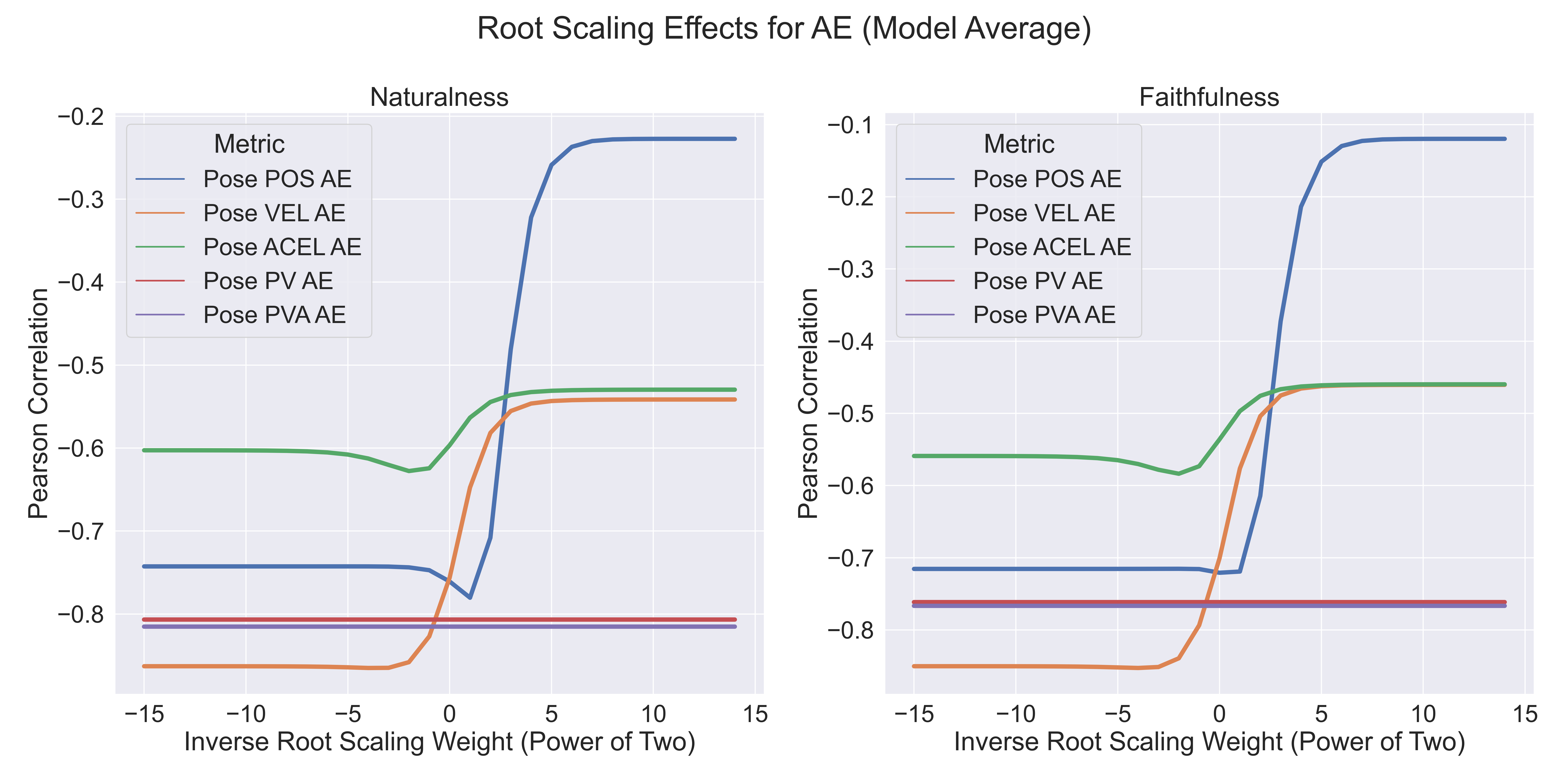}
\includegraphics[width=\linewidth]{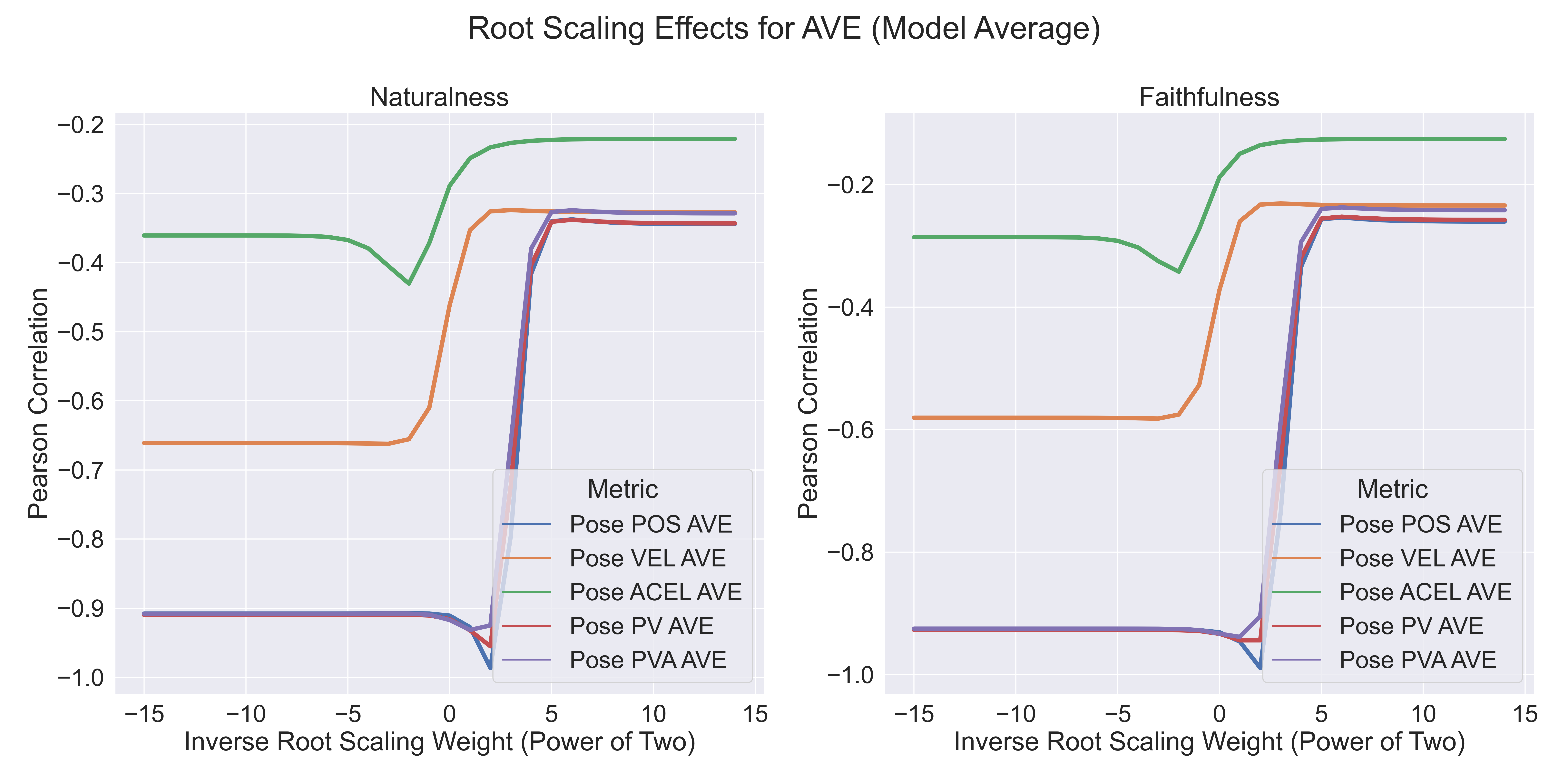}
\caption{Model level examination of root joint scaling effects on CE metrics using all joints. Greater magnitude indicates better performance. }
\label{fig:by-model-root-scale}
\end{figure}

We recommend future evaluations employ our MoBERT evaluator alongside metrics such as R-Precision 1-5, FID, Pose POS AVE, and Root PV AE when assessing text-to-motion generation. Figure \ref{fig:by-model-root-scale} can help determine optimal root scalings for Pose POS AVE.

\subsubsection{MoBERT Out-of-Distribution (OOD) Robustness}

MoBERT was pretrained exclusively on the HumanML3D dataset. Even though the regression versions are trained to fit human judgments using moderately OOD data produced by various generative models, these models were trained to emulate the HumanML3D data. The human judgment fine-tuning potentially learns to harness the most reliable MoBERT output features. These features, inferred from the distinct distributions produced by motion generation models, suggest a potential for MoBERT to withstand OOD scenarios. However, without a substantially OOD dataset, aligned to the 22-joint SMPL body model of HumanML3D, and coupled with human judgments this remains speculative. Low diversity in our training also may result in our vocabulary not being well covered for infrequent tokens. 

To enhance MoBERT's adaptability, future efforts could retrain the regression versions with a growing, diverse dataset of human judgments as they are collected. This could enable MoBERT to better accommodate various motion types, textual inputs, or evolving concepts of \textbf{Naturalness} and \textbf{Faithfulness}. Nonetheless, when adapting MoBERT to OOD data, assessing its performance against relevant human judgments is recommended.
\section{Conclusions}
In this study, we compiled a dataset of human motions generated by recent text-to-motion models, accompanied by human quality assessments. By analyzing existing and newly proposed evaluation metrics, we identified those that best correlate with human judgments. R-Precision is a reliable metric for evaluating model quality, but traditional CE metrics and our novel versions with root and component scaling perform equally well or even better, suggesting that R-Precision should not be relied upon as the sole metric. Some newer metrics that have replaced CE metrics in some publications demonstrated suboptimal or even poor performance. Our novel proposed MoBERT evaluator significantly outperforms all competitors, offering a reliable metric at all levels while being reference free. However, efforts to enhance encoder quality or develop novel metrics to improve sample-level evaluations are further encouraged as well as continued human studies whenever possible.

\subsection{Limitations}

Our dataset with 1400 motion annotations is fairly small for automated evaluation and covers only a small fraction of the HumanML3D test set. Although our study presents strong findings for model-level averages, it includes only five models, making model-level correlations potentially vulnerable to chance. Our interannotator agreement is high, but all human annotation has the potential to introduce biases and noise. We used a single instruction for annotation and alternative instructions might yield different results.

As motion generation techniques continue to advance, the samples used in our study may not accurately represent error distributions in future improved models, potentially affecting the determination of the best metric. Despite the strong correlations observed between some metrics and human judgments, independent human evaluations remain crucial for comparing model performance.

\subsection{Acknowledgements}
This research was partially supported by NSF NRI Grant IIS-1925082 and NSF IIS-2047677 as well as funding from Wormpex AI Research.

\bibliographystyle{ACM-Reference-Format}
\bibliography{bibliography}

\clearpage

\begin{figure}[H]
\centering
\includegraphics[width=\linewidth]{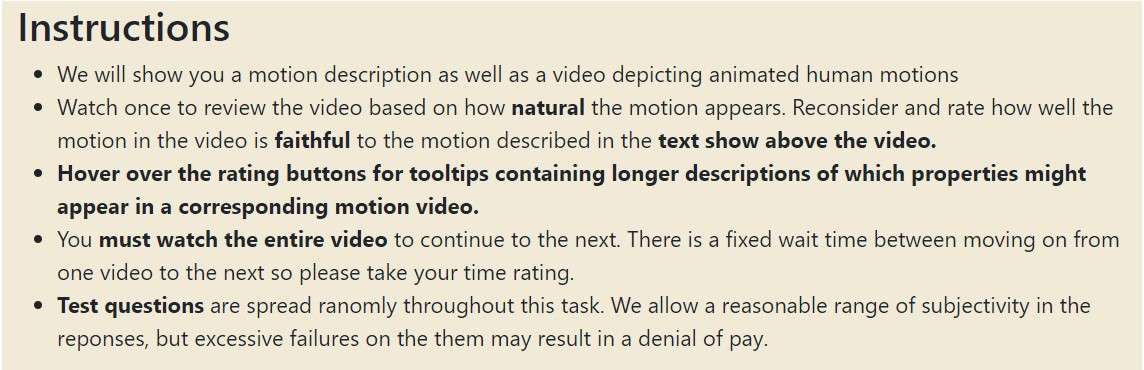}
\caption{Instructions for raters in human judgment evaluations.}
\label{fig:mturk_inst}
\end{figure}

\begin{figure}[H]
\centering
\includegraphics[width=\linewidth]{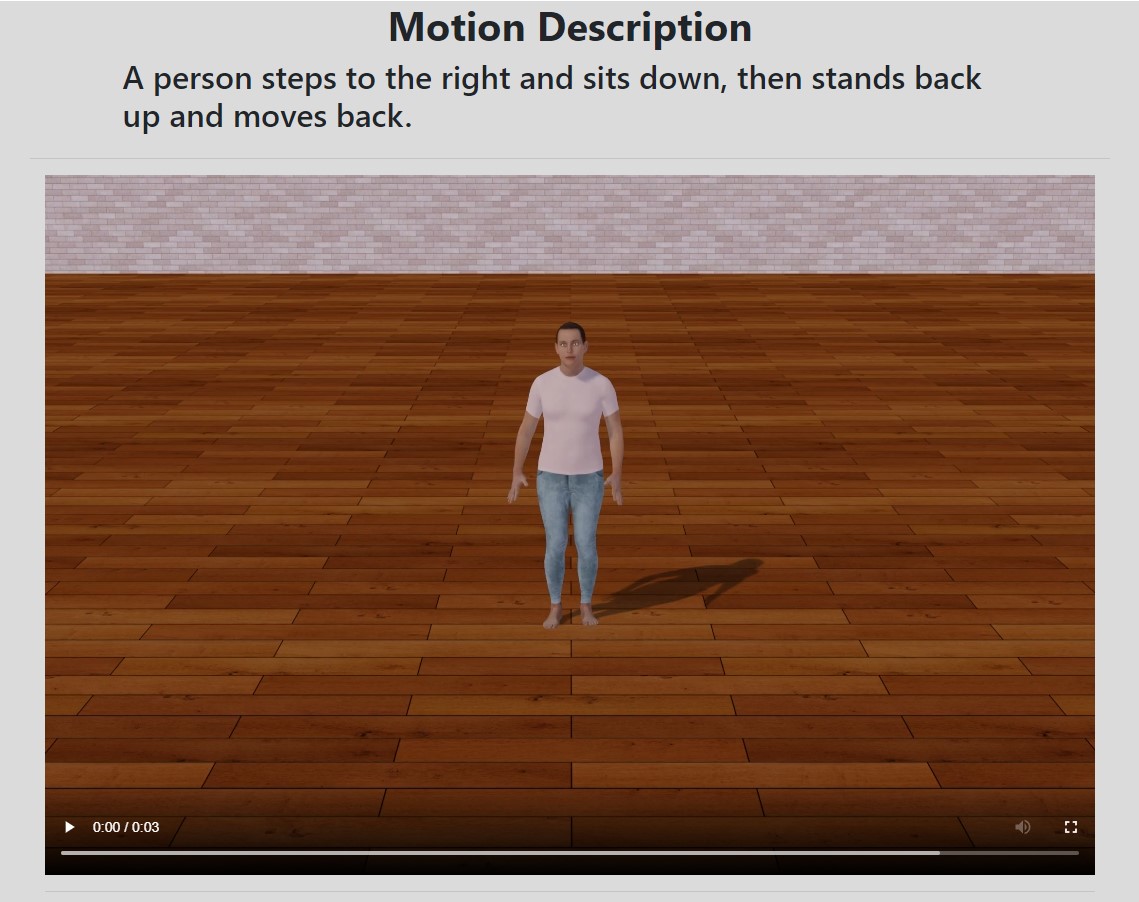}
\caption{UI motion viewing section, situated below the instructions and above the rating selection.}
\label{fig:mturk_view}
\end{figure}

\begin{figure}[H]
\centering
\includegraphics[width=\linewidth]{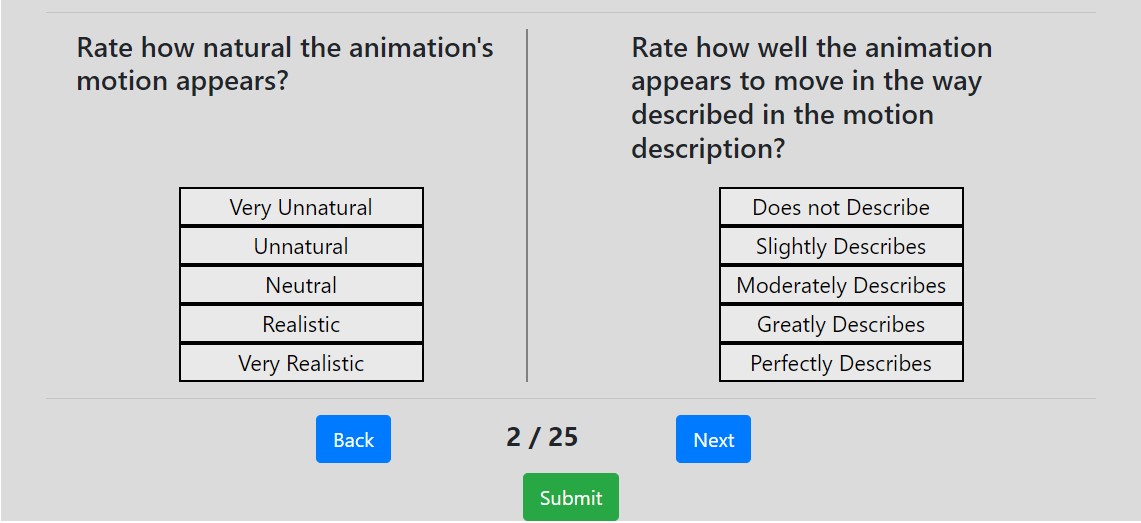}
\caption{Rating selection UI, located below the motion viewing section. Detailed descriptions for each rating option were provided as tooltips upon hovering.}
\label{fig:mturk_rate}
\end{figure}

\begin{table}[H]
\begin{center}
\begin{tabular}{|| c | c ||}
\hline
\multicolumn{2}{|| r ||}{IAA (Krippendorff's Alpha)} \\[0.5ex]
\hline\hline
Naturalness & Faithfulness \\
0.647 & 0.701 \\
\hline
\end{tabular}
\end{center}
\caption{Inter-annotator agreement across all replicated MTurk samples. Results indicate substantial but non-perfect agreement.}
\label{fig:rating-iaa}
\end{table}

\begin{table}[H]
\begin{center}
\begin{tabular}{|| c | c ||}
\hline
\multicolumn{2}{|| c ||}{Pearson's Correlation} \\[0.5ex]
\hline\hline
Sample Level & Model Level \\
0.62 & 0.83 \\
\hline
\end{tabular}
\end{center}
\caption{Correlation between \textit{Naturalness} and \textit{Faithfulness}.}
\label{fig:faith-nat-corr}
\end{table}

\clearpage


\appendix

\end{document}